\documentclass[10pt,journal,compsoc]{IEEEtran}
\usepackage{amssymb}

\usepackage[colorlinks,urlcolor=blue,linkcolor=blue,citecolor=blue]{hyperref}

\usepackage{color,array}

\usepackage{graphicx}
\usepackage{bm}
\usepackage{bbm}
\usepackage{bbding}
\usepackage{epsf}
\usepackage{epsfig}
\usepackage{float}
\usepackage{graphicx}
\usepackage{multirow}
\usepackage{algorithmic}
\usepackage{algorithm}
\usepackage{amsmath}

\newcommand{\etal}{\textit{et al}}
\newcommand{\ie}{\textit{i.e.}}

\newcommand{\revise}[1]{\textcolor{black}{#1}}

\begin{document}

\title{\revise{Adversarial Semantic Augmentation 
for \\ Training Generative Adversarial Networks \\
under Limited Data}}

\author
{Mengping Yang$^{1,2}$, Zhe Wang$^{*,1,2}$, Ziqiu Chi$^{1,2}$, Dongdong Li$^{2}$, Wenli Du$^{1}$
\thanks{$^*$Corresponding author. Email:
        wangzhe@ecust.edu.cn (Z. Wang)
        }
\thanks{$^1$Key Laboratory of Smart Manufacturing in Energy Chemical Process, Ministry of Education, East China University of Science and Technology, Shanghai, 200237, P. R. China.
        }
\thanks{$^2$Department of Computer Science \& Engineering, East China University of Science \& Technology, Shanghai, 200237, P. R. China.
}
}

\maketitle

\begin{abstract}
   \revise{Generative adversarial networks (GANs) have made remarkable achievements in synthesizing images in recent years}.
   Typically, training GANs requires massive data, and the performance of GANs deteriorates significantly when training data is limited. 
   To improve the synthesis performance of GANs in low-data regimes, existing approaches use various data augmentation techniques to enlarge the training sets.
   \revise{However, it is identified that these augmentation techniques may leak or even alter the data distribution}.
   To remedy this, we propose an adversarial semantic augmentation (ASA) technique to enlarge the training data at the semantic level instead of the image level.
   \revise{Concretely, considering semantic features usually encode informative information of images, we estimate the covariance matrices of semantic features for both real and generated images to find meaningful transformation directions}. 
   Such directions translate original features to another semantic representation, \emph{e.g.}, changing the backgrounds or expressions of the human face dataset.
   Moreover, we derive an upper bound of the expected adversarial loss. By optimizing the upper bound, our semantic augmentation is implicitly achieved.
   Such design avoids redundant sampling of the augmented features and introduces negligible computation overhead, making our approach computation efficient.
   \revise{Extensive experiments on both few-shot and large-scale datasets demonstrate that our method consistently improve the synthesis quality under various data regimes, and further visualized and analytic results suggesting satisfactory versatility of our proposed method}.
   %
\end{abstract}



\begin{IEEEkeywords}
Generative Adversarial Networks, Limited Data Generation, Semantic Augmentation
\end{IEEEkeywords}

\section{Introduction}
\label{sec:intro}

\IEEEPARstart{G}{enerative} adversarial networks (GANs)~\cite{goodfellow2014generative, brock2018large, karras2018progressive} have demonstrated conspicuous developments in image synthesis tasks such as image generation~\cite{brock2018large, karras2020analyzing, Karras2021, stylegan3, yang2022wavegan},
domain translation~\cite{zhu2017unpaired, park2019semantic, choi2020stargan}, and image inpainting~\cite{PEPSITNNLS, yu2018generative, Guo2021ICCV}.
Despite the fact that GANs have shown exhilarating success in various applications, training these models is notoriously difficult since it requires a tremendous amount of data and substantial computation resources~\cite{karras2020training}.
Besides, when given limited data (\emph{e.g.}, less than $1,000$), training GANs is more challenging because the data deficiency causes issues like deteriorated synthetic images~\cite{liu2021towards}, memorization of training samples~\cite{karras2020training}, mode collapse~\cite{liu2019spectral}, and training divergence~\cite{DiffAug}.

Recently, few-shot image generation~\cite{liu2021towards, ojha2021fsgan, wang2022fregan, yang2023protogan, yang2023image}, which seeks to stabilize the training and improve the quality of synthesis in the low-data regimes, has drawn increasing attention.
\revise{One intuitive solution for few-shot image generation is to transfer pre-tained knowledge gained from auxiliary domains with sufficient data to target domain with limited data}~\cite{noguchi2019image, mo2020freeze, Wang2020CVPR}.
Such methods, however, still require pre-training on large-scale datasets (\emph{e.g.}, FFHQ with $70,000$ images), are time-consuming and expensive.
\revise{Additionally, the performance degrades when the auxiliary domains and the target domains are disjoint}.
The other straightforward way is to expand the training samples by massive data augmentation techniques~\cite{karras2020training, DiffAug, zhao2020improved}.
Though these recent studies have demonstrated the effectiveness of augmenting training samples, they neglect the fact that conventional data augmentation at the image level (\emph{e.g.}, flipping, rotation) may change the distribution~\cite{tran2021on, karras2020training} and mislead the generator.

\revise{Motivated from the intriguing observations that %
1) deep neural networks are excellent at extracting linearized feature representations~\cite{bengio2013better, upchurch2017deep, ISDATPAMI2021, li2021transferable},
and 2) semantic features in the model's deep feature space usually encode informative representation of images (see Fig.~\ref{fig:activation-obama}), we tackle the data scarcity by proposing an adversarial semantic augmentation (ASA) technique for few-shot image generation.
}
Namely, there are many meaningful semantic transformation directions of one specific sample, and different directions correspond to different semantic transformations.
\revise{After applying these transformations to the semantic features, the obtained semantic representations could be viewed as novel samples in the deep feature space.
Such that, the semantic features could be augmented for more diverse samples along certain transformation directions.
\revise{For instance, the perspective, expression of a person could be altered by transforming the semantic features along corresponding directions}.
Consequently, we augment the features by translating them along with specific semantic transformation directions}.

\begin{figure}
  \centering
  \includegraphics[width=\linewidth]{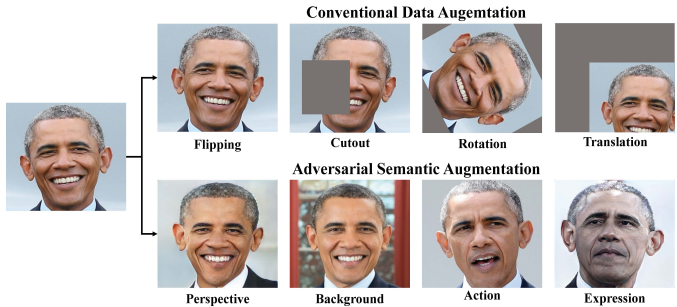}
  \caption{
  \textbf{Comparison between conventional and semantic data augmentation}.
  Conventional and semantic data augmentation perform images at the image and semantic level respectively, making them complementary to each other.
  }
  \label{fig:semantic-augmentation}
  \vspace{-4mm}
\end{figure}

\revise{Fig.~\ref{fig:semantic-augmentation} presents the comparison between conventional and our proposed semantic data augmentation.
Obviously, conventional data augmentation techniques augment images at the image level whereas semantic data augmentation performs image transformation along with meaningful directions, facilitating more diverse augmented images.}
One straightforward way to find such directions is to define transformation directions manually, \emph{e.g.}, define the transformation ``get old", ``laugh" for human-face datasets.
This solution, however, is both computationally intensive and complicated to implement.

\revise{In order to find meaningful directions for semantic transformation, we estimate the covariance matrices of the features for both real and generated samples, which contain possible semantic change directions}.
Moreover, an online updating scheme is designed to calculate the covariance matrices of features to acquire precise estimation~\cite{ISDATPAMI2021}.
Further, we introduce two block attention mechanisms to facilitate the extraction process.
\revise{Such that,  we can augment the samples by sampling from the estimated covariance matrices}.
Instead of augmenting the training samples explicitly for fixed times $S$, we develop an upper bound expected adversarial cross-entropy loss over our adversarial semantic augmentation method.
\revise{By optimizing the upper bound directly, semantic augmentation is applied implicitly and automatically, without requiring adjustment of sampling times manually}. 
As presented in Fig.~\ref{fig:Intro}, our ASA-GAN can generate diverse images with similar semantics but differ in many aspects, \emph{e.g.}, gender, hairstyle, perspective.
We summarize the primary contributions of our method as:

\begin{figure*}
  \centering
  \includegraphics[width=\linewidth]{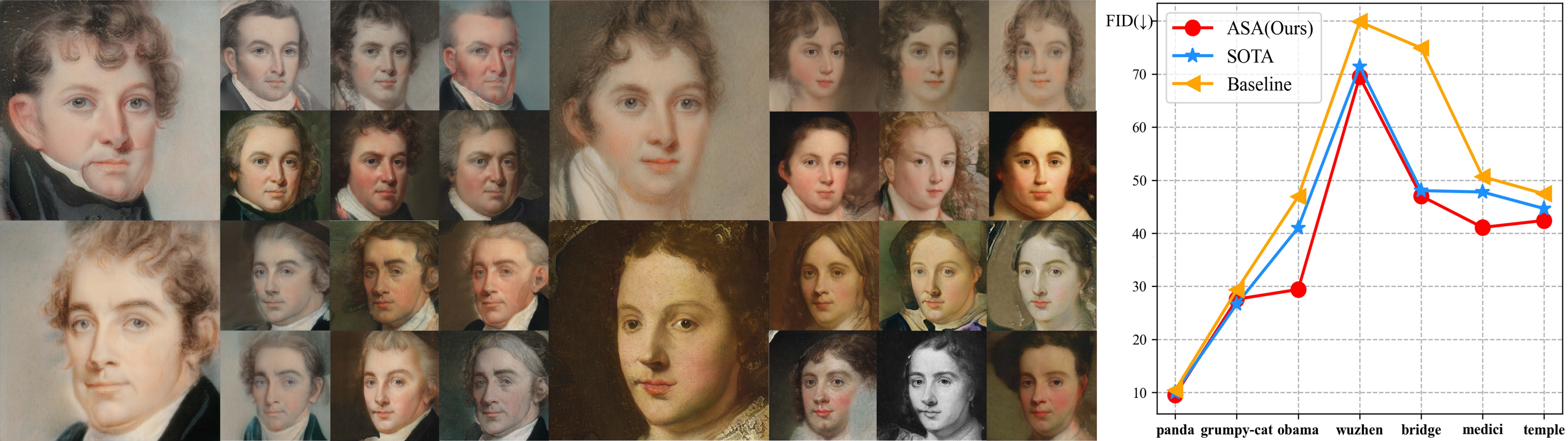}
  \caption{\textbf{Adversarial Semantic Augmentation for Training GANs under Limited Data.}
      (\emph{left}) Synthetic images of high fidelity and diversity produced by our model trained on the limited MetFace~\cite{karras2020training} dataset. These synthetic images share similar semantics but varying details such as age, gender, hairstyle, etc.
      (\emph{right}) FID (lower is better) comparison results of our method with state-of-the-art method FastGAN~\cite{liu2021towards} on 100-shot datasets (with only $100$ training images). }
  \label{fig:Intro}
  \vspace{-4mm}
\end{figure*}

$\bullet$
We propose a novel adversarial semantic augmentation (ASA) approach to facilitate the diversity and fidelity of few-shot image generation. To our knowledge, our ASA is the first work that employs semantic augmentation for GANs.
Notably, our ASA introduces no extra computation costs and makes no change to the backbone of the network, enabling potentials to be plugged into various GANs models.

$\bullet$
We derive an upper bound of the expected discriminant loss. The semantic transformation is performed automatically by optimizing the upper bound.
Moreover, we theoretically prove that our technique maintains the original distribution, providing proper guidelines to the generator.

$\bullet$
We conduct extensive experiments on $20$ few-shot and $3$ large-scale datasets from a wide range of domains. Both qualitative and quantitative results demonstrate the effectiveness of our ASA technique.

\section{Related Work}
In this section, we review approaches most related to our method.
For readers who want to know more about GANs, please refer to~\cite{liu2021generative} and~\cite{Survey2021VAE} for more comprehensive surveys.

\subsection{Few-shot Generative Adversarial Networks}
Training GANs under limited data often results in volatility and overfitting issues, leading to low fidelity and quality of synthesized images.
\revise{Extensive approaches have been proposed to overcome these drawbacks and improve the synthesis quality}.
These techniques can be roughly divided into two categories based on whether pre-trained models are required: 1) approaches based on transfer learning and 2) approaches train from scratch.
\revise{The former assumes that one can facilitate the generalization of the target domain by leveraging the knowledge from the source domain, where massive training data is available for pre-training}.
TransferGAN~\cite{wang2018transferring}, Scale/Shift~\cite{noguchi2019image} and FreezeD~\cite{mo2020freeze} build baselines for adapting knowledge to the target domain by finetuning the pre-trained GANs.
MineGAN~\cite{wang2020minegan} and MineGAN++~\cite{MineGAN++} acquire beneficial information with a well-designed miner network.
ElasticGAN\cite{li2020few} analyzes the weight importance and regularizes the weight changes quantitatively during the adaptive process.
Similarly, FSGAN learns to adapt the pre-trained weights' statistics (singular values) to transfer knowledge~\cite{robb2020few}.
Although these approaches have promoted the fidelity and diversity of GANs in low-data regimes, they still require pre-training on sufficient data.
\revise{Besides, when the domain shift between the source and target domain is large, the performance degrades significantly due to negative transfer}.

Approaches directly trained on limited data alleviate the overfitting problem by adopting regularization or data augmentation techniques.
Liu~\etal~\cite{liu2021towards} design FastGAN with a skip-layer excitation (SLE) block to fuse low-resolution information into high-resolution feature maps. They further propose a self-supervised discriminator as a regularization to reconstruct training images.
LeCam-GAN improves the performance and stability by adding a regularized loss~\cite{tseng2021regularizing}.
Differently, Projected-GAN~\cite{projectedGAN} and Aided-GAN~\cite{kumari2022ensembling} improve the synthesis performance by leveraging off-the-shelf discriminative models as discriminators.
Other approaches employ various data augmentation techniques to increase training samples, we briefly review them in~\ref{Sec:RelatedDA}.

\subsection{Data Augmentation in GANs}
\label{Sec:RelatedDA}
Inspired by the great success of data augmentation in training discriminant models~\cite{zhang2018mixup, shorten2019survey}, researchers have investigated the effectiveness of applying data augmentation in training GANs recently.
Zhao~\etal~~\cite{zhao2020image} provide guidelines on augmenting training samples for both vanilla GAN and GANs with improved techniques such as consistency regularization~\cite{CR2020}.
Borrowing the consistency principle from semi-supervised learning, CR-GAN penalizes the sensitivity of the discriminator to the augmentations on real images~\cite{CR2020}, and ICR-GAN further improves the performance by augmenting both real and generated images~\cite{zhao2020improved}.
Zhao~\etal~~\cite{DiffAug} design a differentiable augmentation approach to stabilize training, and Karras~\etal~~\cite{karras2020training} perform an adaptive strategy to control the strength of augmentation.
ContraD\cite{jeong2021training} and InsGEN~\cite{yang2021insgen} train GANs with strong augmentations in a self-supervised way to improve the representation ability of $D$.
Theoretically, Tran~\etal~\cite{tran2021on} prove that classical DA (\emph{e.g.}, rotation, cropping) may mislead the generator and propose a DAG framework for performing data augmentation in GANs.
\revise{Unlike prior approaches that augment images in the original image level, we enlarge the training sets implicitly in semantic feature space, without altering the data distribution and produce representative semantic features for augmentation}.

\section{Proposed Approach}

\subsection{Preliminaries}
\label{motivation}
A GAN model is typically formulated as a min-max game between a generator $G$ and a discriminator $D$.
The former tries to approach real data distribution by adversarial learning against the latter, which attempts to distinguish real samples from generated ones.
The training scheme of the GAN models can be formally expressed as:
\begin{equation}
\max _{D} \mathcal{L}_{D}, \mathcal{L}_{D}=\underset{\boldsymbol{x} \sim T}{\mathbb{E}}\left[f_{D}(D(\boldsymbol{x}))\right]+\underset{\boldsymbol{z} \sim Z}{\mathbb{E}}\left[f_{G}(D(G(\boldsymbol{z})))\right],
\end{equation}



\begin{equation}
\min _{G} \mathcal{L}_{G}, \mathcal{L}_{G}=\underset{\boldsymbol{z} \sim Z}{\mathbb{E}}\left[h_{G}(D(G(\boldsymbol{z})))\right],
\end{equation}
\revise{where $Z$ denotes the prior distribution (\emph{e.g.}, Gaussian distribution) of the latent code $Z$ and $T$ denotes the distribution of the training data}.
The notations $f_{D}$, $f_{G}$ and $h_{G}$ are mapping functions that process the semantic features of input images, and various losses of GANs like hinge version loss~\cite{heusel2017gans} and binary cross-entropy loss~\cite{radford2016unsupervised} can be derived from the mapping functions.
We regard the process of the discriminator $D$ distinguishing real samples from generated ones as a binary classification problem.
\revise{The min-max loss is defined as}:

\begin{equation}
\min _{G} \max _{D} V(D, G)=E_{x \sim T} [\log D(x)]+E_{z \sim Z}[\log (1-D(G(z)))].
\end{equation}


\subsection{Adversarial Semantic Augmentation}

\begin{figure}
  \centering
  \includegraphics[width=\linewidth]{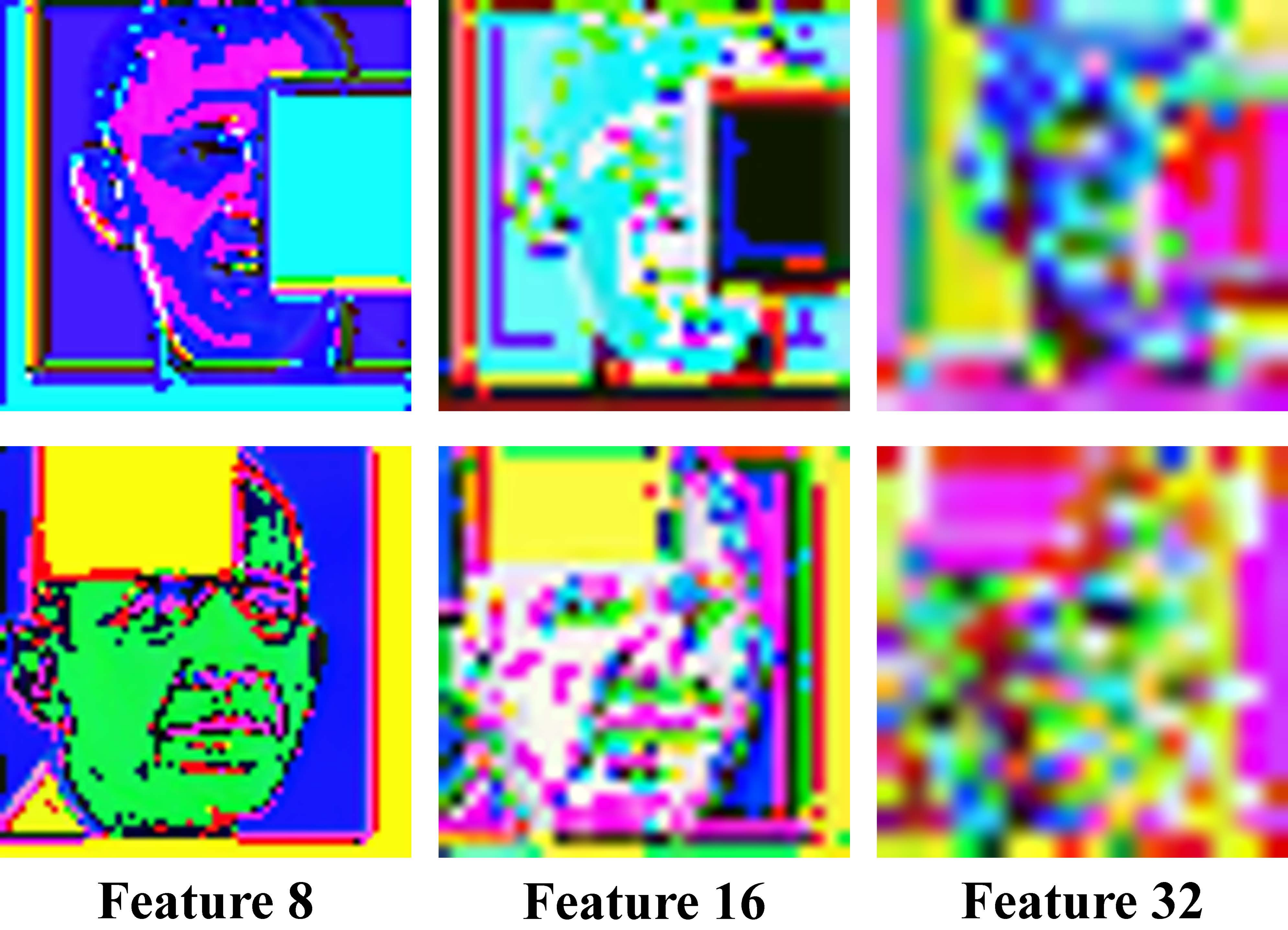}
  \caption{
  \revise{\textbf{Activation from different layers of the network on the 100-shot-Obama dataset}.
  Different distributional information is captured at different semantic level, and the semantic features goes from general to abstract as the network goes deeper.
  }}
  \label{fig:activation-obama}
  \vspace{-4mm}
\end{figure}

\begin{figure*}
  \centering
  \includegraphics[width=\linewidth]{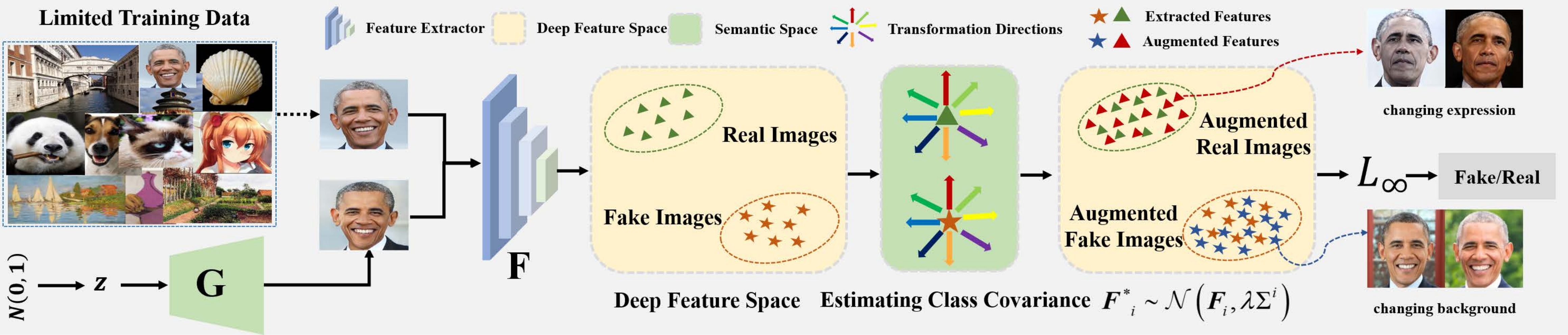}
  \caption{\textbf{\revise{The overall framework of our proposed adversarial semantic augmentation(ASA) model.}}
  We augment the training data in the semantic space by translating the features of both real and fake images along meaningful transformation directions.
  Such directions are obtained by estimating the covariance matrices for both real and fake images.
  ASA-GAN performs semantic augmentation implicitly by optimizing an upper bound of adversarial loss, which is computation efficient and easy-to-implement.
  }
  \label{Framework}
  \vspace{-4mm}
\end{figure*}

\revise{
Fig.~\ref{fig:activation-obama} presents the activation maps of different semantic features encoded by various layers of the network.
Obviously, various layers of the model capture different semantic features of training images at different semantic level, and the features get more abstract as the network goes deeper (\emph{i.e.,} from left to right).
These visualization results reflect the intuitive idea of our proposed adversarial semantic augmentation. 
Specifically, when translating these features to certain semantic direction, novel features could be obtained for augmenting the training sets in the semantic level.
}

\revise{Fig.~\ref{Framework} presents the overall pipeline of our proposed adversarial semantic augmentation for training GANs under limited data}.
\revise{Considering that directly defining or finding meaningful transformation directions is nontrivial with innumerable directions in the feature space, and annotating such directions requires massive resources.
Alternatively, we estimate the intra-class covariance matrices for both real (\emph{i.e.,} $\Sigma^{(r)}$)  and generated images (\emph{i.e.,} $\Sigma^{(f)}$) in the deep feature space.
The covariance matrices capture the variation of samples and contain possibly rich directions for transformation.
Accordingly, we augment the deep features by sampling from the estimated covariance matrix $\mathcal{N}\left(0, \Sigma\right)$, which captures rich semantic information.}
Equivalently, the augmented semantic feature $F^{*}_{i}$ can be expressed as: $F^{*}_{i} \sim \mathcal{N}\left(F_{i}, \lambda \Sigma^{{i}}\right)$, where $\Sigma^{{i}}$ denotes the estimated covariance matrices and $\lambda$ is a hyperparameter to control the strength of semantic augmentation.
We set $\lambda=(t / T)$ in implementation, where $T$ is the total iteration and $t$ is the current iteration.
Intuitively, we sample from the estimated covariance matrices for S times and obtain augmented features as ${F^{*}_{i}=[{f^{*}}_{i1},{f^{*}}_{i2},...,{f^{*}}_{iS}]}$.
The discriminator $D$ distinguishes the generated images from real ones by minimizing the binary cross-entropy loss:


\begin{equation}
\mathcal{L}_{S}(\boldsymbol{W}, \boldsymbol{b}, \boldsymbol{\Theta})=\frac{1}{N} \sum_{i=1}^{N} \frac{1}{S} \sum_{k=1}^{S}-\log \left(\frac{e^{\boldsymbol{w}_{\tilde{y_{i}}}^{T} \boldsymbol{f}_{i}^{k}+\boldsymbol{b}_{\tilde{y_{i}}}}}{\sum_{j=1}^{C} e^{\boldsymbol{w}_{j}^{T} \boldsymbol{f}_{i}^{k}+\boldsymbol{b}_{j}}}\right),
\end{equation}
\revise{where $N$ denotes the number of samples and $S$ represents the number of augmentation times}. 
$\tilde{y_{i}}$ is the pseudo label and $C$ is the total number of classes, which equals $2$ in our model.
$\boldsymbol{W}$ and $\boldsymbol{b}$ denotes the weights and biases of the classification layer of the discriminator, respectively.

\revise{By increasing the augmentation times $S$, more semantic features could be obtained.
However, the extra computation cost is non-negligible when $S$ is large and the semantic features in deep feature space have high dimensions}.
Instead of explicitly augmenting the features for $S$ times, we consider to augment the features for infinite times, \ie, $S\to+\infty$:
%

\revise{
\begin{equation}
\begin{scriptsize}
\begin{aligned}
    \label{infinity}
    \mathcal{L}_{S\to+\infty}(\boldsymbol{W}, \boldsymbol{b}, \boldsymbol{\Theta})&=\frac{1}{N} \sum_{i=1}^{N} \mathrm{E}_{\boldsymbol{{f^{*}}}_{i}}\left[-\log \left(\frac{e^{\boldsymbol{w}_{\tilde{y_{i}}}^{T} \boldsymbol{f^{*}}_{i}+b_{\tilde{y_{i}}}}}{\sum_{j=1}^{C} e^{\boldsymbol{w}_{j}^{\mathrm{T}} \boldsymbol{f^{*}}_{i}+b_{j}}}\right)\right] \\
    &=\frac{1}{N} \sum_{i=1}^{N} \mathrm{E}_{\boldsymbol{{f^{*}}}_{i}}\left[\log \left(\sum_{j=1}^{C} e^{\left(\boldsymbol{w}_{j}^{T}-\boldsymbol{w}_{\tilde{y_{i}}}^{T}\right) \boldsymbol{f^{*}}_{i}+\left(b_{j}-b_{\tilde{y_{i}}}\right)}\right)\right].
\end{aligned}
\end{scriptsize}
\end{equation}
}

Eq.~\ref{infinity} is not computation efficient as $\mathrm{E}_{\boldsymbol{{f^{*}}}_{i}}$ can not be calculated precisely.
Alternatively, we minimize $\mathcal{L}_{S\to+\infty} $ by deriving an upper bound of Eq.~\ref{infinity}.
Following Jensen's inequality and the properties of convex function $\log (x)$, we have $\mathrm{E}[\log X] \leq \log \mathrm{E}[X]$, by substituting Eq.~\ref{infinity} into the Jesen's inequality, we can obtain:


\begin{equation}
\footnotesize
\begin{aligned}
\label{upperbound}
&\mathcal{L}_{S \rightarrow+\infty}(\boldsymbol{W}, \boldsymbol{b}, \boldsymbol{\Theta})  \\
&\leq \frac{1}{N} \sum_{i=1}^{N} \log \left(\sum_{j=1}^{C} \mathrm{E}_{\boldsymbol{{f^{*}}}_{i}}\left[e^{\left(\boldsymbol{w}_{j}^{T}-\boldsymbol{w}_{y_{i}}^{T}\right) \boldsymbol{{f^{*}}}_{i}+\left(\boldsymbol{b}_{j}-\boldsymbol{b}_{y_{i}}\right)}\right]\right) \\
&=\!\frac{1}{N}\! \sum_{i=1}^{N} \log\! \left(\sum_{j=1}^{C} e^{\left((\boldsymbol{w}_{j}^{T}-\boldsymbol{w}_{y_{i}}^{T}) \boldsymbol{{f^{*}}}_{i}+\left(\boldsymbol{b}_{j}-\boldsymbol{b}_{y_{i}}\right)+\frac{\lambda}{2}\left(\boldsymbol{w}_{j}^{T}-\boldsymbol{w}_{y_{i}}^{T}\right) \Sigma^{i}\left(\boldsymbol{w}_{j}-\boldsymbol{w}_{\tilde{y}_{i}}\right)\right.\!}\right) \\
&=\overline{\mathcal{L}}_{\infty}.
\end{aligned}
\end{equation}

$\overline{\mathcal{L}}_{\infty}$ is the derived upper bound of $\mathcal{L}_{S\to+\infty}$.
\revise{In this way, by directly optimizing $\overline{\mathcal{L}}_{\infty}$, we implicitly perform augmentation without manually adjusting the hyper-parameter augmentation times $S$, which is efficient and easy to implement.
When $\lambda=0$, Eq.~\ref{upperbound} degenerates into the conventional classification loss, where no features are semantically augmented}.

Considering the input images of the discriminator are either real or generated, and there are no labels available, we regard the process of distinguishing real and generated images as a binary classification problem in implementation.
The pseudo labels of generated images and real images are set as vectors of $\boldsymbol{0}_s$ and $\boldsymbol{1}_s$ with the size of the batch size, respectively.

\revise{
\begin{equation}
\begin{aligned}
\label{ourloss}
{\overline {\mathcal{L}} _\infty } = \sum\limits_{{\rm{i = }}1}^{\rm{N}} \begin{array}{l}
\log [{{\rm{e}}^{((w^T - w_f^T){{\bf{f}}_f} + ({b} - {b_f}) + \frac{\lambda }{2}(w^T - w_f^T)\mathop \sum \limits^f ({w} - {w_f}))}}\\
 + {{\rm{e}}^{((w^T - w_r^T){{\bf{f}}_r} + ({b} - {b_r}) + \frac{\lambda }{2}(w^T - w_r^T)\mathop \sum \limits^r ({w} - {w_r}))}}]
\end{array} ,
\end{aligned}
\end{equation}
}
\revise{
where the subscript $r$ and $f$ denote the real and synthetic images, respectively.
$w$, $w_f$, and $w_r$ represent the weight of the classification layer of the discriminator correspond the the whole samples, the generated images, and the real images respectively.
By minimizing Eq.~\ref{ourloss}, we augment the semantic features for infinite times implicitly.
}

\textbf{Proof.} Eq.~\ref{upperbound} is derived from the moment-generating function:
\begin{equation}
\begin{aligned}
\label{Eq:moment}
\mathrm{E}\left[e^{t X}\right]=e^{t \mu+\frac{1}{2} \sigma^{2} t^{2}}, \quad X \sim \mathcal{N}\left(\mu, \sigma^{2}\right),
\end{aligned}
\end{equation}
note that $F^{*}_{i} \sim \mathcal{N}\left(\boldsymbol{F}_{i}, \lambda \Sigma^{{i}}\right)$, according to the properties of the normal distribution, we have:

\begin{equation}
\footnotesize
\begin{aligned}
\label{momentgenerate}
&\left(\mathbf{w}_{j}^{T}-\mathbf{w}_{\tilde{y}_{i}}^{T}\right) \mathbf{F}_{i}^{*}+\left(b_{j}-b_{\tilde{y_{i}}}\right) \notag \sim\\
& \mathcal{N}\left(\left(\mathbf{w}_{j}^{T}-\mathbf{w}_{\tilde{y}_{i}}^{T}\right) \mathbf{f}_{i}+\left(b_{j}-b_{\widetilde{y}_{i}}\right), \lambda\left(\mathbf{w}_{j}^{T}-\mathbf{w}_{\tilde{y}_{i}}^{T}\right) \Sigma^{i}\left(\mathbf{w}_{j}-\mathbf{w}_{\tilde{y}_{i}}\right)\right).
\end{aligned}
\end{equation}

By substituting Eq.~\ref{momentgenerate} into Eq.~\ref{Eq:moment} with $t=1$, $\mu=\left(\boldsymbol{w}_{j}^{T}-\boldsymbol{w}_{\tilde{y}_{i}}^{T}\right) \boldsymbol{f}_{i}+\left(b_{j}-b_{\tilde{y}_{i}}\right)$ and $\sigma=\lambda\left(\boldsymbol{w}_{j}^{T}-\boldsymbol{w}_{\tilde{y_{i}}}^{T}\right) \Sigma^{i}\left(\boldsymbol{w}_{j}-\boldsymbol{w}_{\tilde{y_{i}}}\right)$, we obtain $\overline{\mathcal{L}}_{\infty}$ in Eq.~\ref{upperbound}.

\subsection{Online Estimation of Class Covariance matrices}
\label{sec:DynamicEstimation}
To calculate the covariance matrices of all the features precisely, we employ an online approach to aggregate all the statistics of each mini-batch \cite{ISDATPAMI2021}:
\begin{equation}
\label{mu}
{\mu}_{j}^{(t)}=\frac{n_{j}^{(t-1)} {\mu}_{j}^{(t-1)}+m_{j}^{(t)} {\mu}_{j}^{\prime(t)}}{n_{j}^{(t)}},
\end{equation}

\revise{\begin{align}
\footnotesize
\label{sigma}
\Sigma_{j}^{(t)}&=
\frac{n_{j}^{(t-1)} \Sigma_{j}^{(t-1)}+m_{j}^{(t)} \Sigma_{j}^{\prime(t)}}{n_{j}^{(t)}} \notag\\
&+ \frac{n_{j}^{(t-1)} m_{j}^{(t)}\left({\mu}_{j}^{(t-1)}-{\mu}_{j}^{(t)}\right)\left({\mu}_{j}^{(t-1)}-{{\mu}}_{j}^{\prime(t)}\right)^{T}}{{n_{j}^{(t)}}^{2}},
\end{align}
}
where $n_{j}^{(t)}=n_{j}^{(t-1)}+m_{j}^{(t)}$ denotes the total number of input images and $m_{j}^{(t)}$ is the number of training samples of every batch, which we set as $8$ in implementation. The subscript $j$ denotes the class label of images in the original paper~\cite{ISDATPAMI2021}, since our model is totally unsupervised and only real and generated images are estimated in our GAN model, the subscript $j$ denotes the real and generated images.
$\mu_{j}^{(t)}$ and ${\mu}_{j}^{\prime(t)}$ denotes the average values of the features at the $t^{th}$ iteration step and the $t^{th}$ batch of samples, respectively.
Similarly, $\Sigma_{j}^{(t)}$ and ${\Sigma}_{j}^{\prime(t)}$ denotes the estimated covariance matrices of the features at the $t^{th}$ iteration step and the $t^{th}$ batch of samples, respectively.

\noindent{\textbf{Computational complexity}.}
\revise{As discussed above, the proposed adversarial semantic augmentation (ASA) is accomplished implicitly by directly optimizing the derived upper bound}.
Consequently, ASA only requires additional computation resources to estimate the covariance matrices and minimize the derived loss in Eq.~\ref{upperbound}.
For a batch of input images, estimating the covariance matrices requires $\mathcal{O}(D^2)$ computational resources, where $D$ is the semantic features' dimension.
Moreover, the computational complexity of optimizing the upper bound loss is $\mathcal{O}(B \times D^2)$, where $B$ denotes the batch-size.
Typically, a commonly used convolutional operation requires $\mathcal{O}(D^2 \times K^2 \times C_{in} \times C_{out})$, where $K$ represents kernel size.
$C_{in}$ and $C_{out}$ denote the input and output channel, respectively.
\revise{By contrast, the computational complexity of our proposed ASA technique can be ignored, enabling efficient semantic augmentation}.

\subsection{Jesen-Shannon(JS) Preserving with Semantic Augmentation}
%
\revise{Conventional data augmentation which performed on the image level typically increases the size of the datasets by applying content-preserving transformations like rotation and cropping to the input images.}
However, applying such transformation $\mathcal{T}$ may change the original distribution of the training set ($P_{d}^{\mathcal{T}} \neq P_{d} $).
When trained on the augmented dataset, the generator is trained to approach the distribution of augmented data via minimizing the JS divergence between its distribution $P_{g}$ and that of augmented dataset $P_{d}^{\mathcal{T}}$ as following (please refer to~\cite{goodfellow2014generative} for detailed derivation):

\revise{
\begin{figure*}
  \centering
  \includegraphics[width=\linewidth]{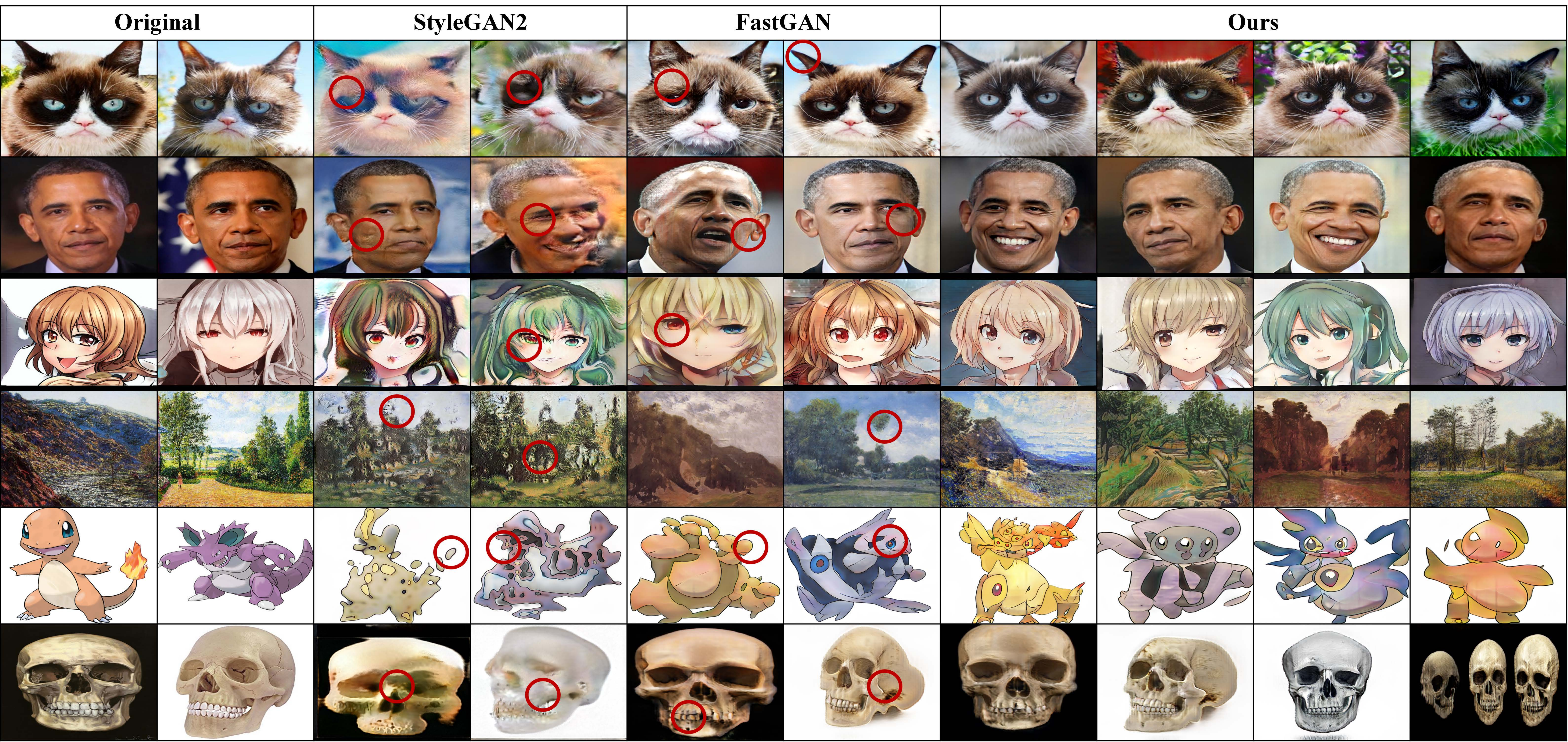}
  \caption{\revise{
  \textbf{Qualitative comparison between our approach and current state-of-the-art models} StyleGAN2~\cite{karras2020analyzing} and FastGAN~\cite{liu2021towards}. The images are synthesized by the saved checkpoints with the lowest FID corresponding to each model.
    The red circles highlight areas with obvious artifacts and distortions.
    Zoom in for a better view.
  }
  }
  \label{VisAll}
  \vspace{-5mm}
\end{figure*}
}

\vspace{-3mm}
\begin{equation}
\begin{aligned}
\label{JS}
\mathcal{V}\left(D^{*}, G\right)=-\log (4)+2 \cdot \mathrm{JS}\left(P_{d}^{\mathcal{T}} || P_{g}\right).
\end{aligned}
\end{equation}
\vspace{-4mm}

Tran \emph{et al.} ~\cite{tran2021on} proved that when and only when the applied augmentations are invertible, \revise{the discriminator $D$ is trained to approach the same distribution as the original distribution $P_{d}$, or the distribution of the original data is changed with nonreversible augmentation}.
Our proposed adversarial semantic augmentation (ASA) implicitly increases the number of semantic features of images without any change to the original data.
Our estimated covariance matrices of images are computed from the statistics of the features of training samples, which is consistent with the original distribution $P_{d}$.
The consistent optimize objective suggests that the generator of our ASA-GAN attempts to minimize the JS divergence between the generated distribution $P_{g}$ and the original distribution $P_{d}$.
Therefore, our augmentation technique ASA makes no change to the original distribution.
The augmented semantic features are leveraged to help the generator $G$ to synthesize diverse and plausible images, especially when given limited training data.

\subsection{Enhancing the Feature Representation}

The feature extractor of our discriminator is based on CNNs. Thus we adopt two efficient and effective attention mechanisms~\cite{woo2018cbam}, \ie, channel attention and spatial attention, to increase the representative ability of the feature extracting process.

The channel attention operates features along with channel axes to learn ``what'' is essential by aggregating spatial information using average and max pooling:

\vspace{-3mm}
\begin{equation}
\begin{aligned}
\mathbf{A}_{\mathbf{c}}(\mathbf{f})=\sigma\left(\mathbf{W}_{\mathbf{1}}\left(\mathbf{W}_{\mathbf{0}}\left(\mathbf{f}_{\mathbf{a v g}}^{\mathbf{c}}\right)\right)+\mathbf{W}_{\mathbf{1}}\left(\mathbf{W}_{\mathbf{0}}\left(\mathbf{f}_{\mathbf{max}}^{\mathbf{c}}\right)\right)\right),
\end{aligned}
\end{equation} where $\sigma(*)$ denotes the sigmoid activation function, $\mathbf{W}_{\mathbf{0}}$ and $\mathbf{W}_{\mathbf{1}}$ is the weight of MLP layer, the BatchNorm and Relu activation function are followed by $\mathbf{W}_{\mathbf{0}}$ in our implementation.

The spatial attention operates features along with spatial axes to learn ``where'' is necessary. The spacial attention is illustrated as:

\vspace{-4mm}
\begin{equation}
\begin{aligned}
\mathbf{A}_{\mathbf{s}}(\mathbf{f})=\sigma\left(\mathbf{conv^{k \times k}}([\mathbf{f}_{\mathbf{avg}}^{\mathbf{s}},\mathbf{f}_{\mathbf{max}}^{\mathbf{s}}])\right),
\end{aligned}
\end{equation}
where $([*])$ denotes the concatenate operation, and $conv^{\mathbf{k \times k}}$ is the convolutional operation with the filter size of $\mathbf{k \times k}$.
By leveraging the extracted informative features, we estimate the covariances of features more precisely.

\subsection{Learning Objective}
\label{Objective}
The feature extractor of our discriminator $D$ encodes training samples into semantic feature space.
We treat $D$ as an encoder and train small decoders to reconstruct the training samples with a reconstruction loss following~\cite{liu2021towards}, which enhances the ability of $D$ to extract features.

\begin{equation}
\begin{aligned}
\mathcal{L}_{recons}=\mathbb{E}_{\mathbf{f} \sim D(x), x \sim I_{train}}[||\mathcal{G}(\mathbf{f})-\mathcal{T}(x)||],
\end{aligned}
\end{equation}
where $x$ denotes the training samples and $f$ is the extracted features from $D$. $\mathcal{G}$ and $\mathcal{T}$ are operations on the features and training images, respectively.

In sum, the learning objective of our ASA-GAN is given as follows:

\vspace{-3mm}
\begin{equation}
\begin{aligned}
\label{eq:discriminator}
{{\cal L}_D} = \min ({\overline {\cal L} _\infty }{\rm{ + }}{ \cal L} _{recons}),
\end{aligned}
\end{equation}

\vspace{-5mm}
\begin{equation}
\begin{aligned}
\label{eq:generator}
{{\cal L}_G} = \max {\overline {\cal L} _\infty } = \min ({- \overline {\cal L} _\infty}).
\end{aligned}
\end{equation}
\vspace{-5mm}

When other versions of the adversarial losses (\emph{e.g.}, Hinge version loss~\cite{lim2017geometric}) are employed, our proposed technique can be easily implemented into other GAN's frameworks as an effective and efficient regularization item.
\revise{Algorithm~\ref{alg} presents The pseudocode of our proposed ASA-GAN, which provides a clear understanding of our method}.
\begin{center}
    \vskip -0.1in
        \begin{algorithm}[H]
            \caption{The pseudocode of our ASA-GAN algorithm.}
            \label{alg}
        \begin{algorithmic}[1]
            \STATE {\bfseries Input:} Training images $\mathcal{D}$, iteration $T$
            \STATE Randomly initialize parameters $\bm{W}, \bm{b}$ and $\bm{\Theta}$
            \FOR{$t=0$ {\bfseries to} $T$}
            \STATE Sample a mini-batch $\{ \bm{x}_i \}_{i=1}^B$  from $\mathcal{D}$
            \STATE Sample a mini-batch $\{ \bm{z}_i \}_{i=1}^B$ from $Z \in \mathcal{N}(0,1)$
            \STATE Generate a mini-batch images with the generator $G$
            \STATE Estimate the covariance matrices of real and generated images (Eq.~\ref{mu} and Eq.~\ref{sigma})
            \STATE Compute $\overline L _\infty$ in Eq.~\ref{ourloss}
            \STATE Update the discriminator by optimizing Eq.~\ref{eq:discriminator}
            \STATE Update the generator by optimizing Eq.~\ref{eq:generator}
            \ENDFOR
            \STATE {\bfseries Output:} model parameters $\bm{W}, \bm{b}$ and $\bm{\Theta}$
        \end{algorithmic}
        \end{algorithm}
\end{center}

\section{Experiments}
\subsection{Experiments Setups}
\textbf{Datasets.}
We investigate the performance of our approach by performing comprehensive experiments on multiple datasets from a range of categories, including realistic photos, human faces, art-like and anime-style images.
On 256*256 resolution, we evaluate on Animal-Face~\cite{si2011learning} and 100-shot datasets~\cite{DiffAug}.
On 512*512 resolution, we evaluate on Anime Face, Art Paintings, Moongate, Flat-colored, and Fauvism-still-life datasets~\cite{liu2021towards}.
On 1024*1024 resolution, we evaluate on Pokemon, Skulls, Shells, BrecaHAD~\cite{aksac2019brecahad}, MetFace~\cite{karras2020training} and Flowers~\cite{nilsback2006visual}.
For those datasets that are close but not exactly the specific resolution like 512*512, we resize them to the closest resolution in implementation.

\textbf{Evaluated Metrics.}
We adopt two common evaluation metrics: Fr\'{e}chet Inception Distance (FID)~\cite{FIDnips2017} and Inception Score (IS)~\cite{salimans2016improved} to measure the quality of the synthesized images.
Excepted as otherwise noted, we evaluate our model and comparison methods under the same settings following~\cite{liu2021towards} and~\cite{DiffAug}.

\textbf{Baseline and Compared Methods.} The baseline model is integrated from various techniques that have been demonstrated effective for training GANs, namely Spectral Normalization (SN)~\cite{miyato2018spectral}, SLE, and self-supervised discriminator~\cite{liu2021towards}, and we use DiffAug~\cite{DiffAug} following~\cite{liu2021towards}.
We use the binary cross-entropy loss for baseline, which is the special case of Eq.~\ref{ourloss} when $\lambda=0$.

We compare our model with three types of few-shot GANs. The first group of methods transfer knowledge from pre-trained models on auxiliary domains, namely Scale/shift~\cite{noguchi2019image}, FreezeD~\cite{mo2020freeze}, MineGAN~\cite{wang2020minegan}, TransferGAN~\cite{wang2018transferring}.
Second, we compare with the state-of-the-art model on unsupervised image synthesize StyleGAN2~\cite{karras2020analyzing}.
Finally, we compare with data augmentation based approaches differentiable data augmentation (DiffAug)~\cite{DiffAug}, adaptive data augmentation (ADA)~\cite{karras2020training} and architecture variants model FastGAN~\cite{liu2021towards}.

\textbf{Implementation Details.} We implement our model using the PyTorch framework and evaluate it on a computer equipped with an Intel(R) Xeon(R) Gold 5218 CPU running at @ 2.30GHz and a Nvidia GeForce 3090 GPU(*1).
The batch size is set to $8$ and the total number of iterations is set to $100,000$.
We save the checkpoints every $10K$ iterations and use the saved checkpoints to synthesize images for evaluation.
The learning rate is set as $2e^{-4}$ and the Adam optimizer~\cite{kingma2014adam} is used with ($\beta1$, $\beta2$)= (0.5, 0.999).

We use the trained model of each method to generate $5K$ images for evaluation and comparison under consistent conditions.
For FID, following \cite{liu2021towards} and \cite{DiffAug}, we utilize the entire training images as the reference distribution to calculate the divergence between the distribution of the real images and that of the generated images.
For IS, we use those checkpoints corresponding to the best FID to synthesize $5K$ images, and we devide the synthesized images into $10$ parts to compute the average and standard deviation values of IS.
For every evaluation, we test our model three times and report the best results. On average, the relative accuracy is greater than $95$ percent.
Our code and models will be released upon acceptance.

\subsection{Main Results}
\textbf{Semantic Augmentation at different level.}
We employ ASA under two conditions to investigate the influence of applying our ASA technique at different levels: augment the discriminator ($D$) only and increase both $D$ and the generator ($G$).
The results in Tab.~\ref{Table:Augment} illustrate that better results are obtained when both $D$ and $G$ are augmented, which motivates us to perform our ASA technique on both $D$ and $G$ in other experiments.
Such observation is reasonable since augmenting both the real and synthesized images facilitates the discriminator's representation ability, leading to the discriminator providing effective feedback to the generator.
\begin{table}[htbp]
\caption{\textbf{FID({$\downarrow$}) and IS({$\uparrow$}) comparison on augmenting} neither $D$ nor $G$, augmenting $D$ only, and augmenting both $D$ and $G$.}
\label{Table:Augment}
\resizebox{.5\textwidth}{!}{
\begin{tabular}{llcccccc}
\hline
\multicolumn{8}{c}{\textbf{256*256 Datasets}}                                                                                                                                         \\ \hline
\multicolumn{2}{l}{Augment?}                  & \multicolumn{2}{c}{AnimalFace-cat}      & \multicolumn{2}{c}{100-Shot-Obama}               & \multicolumn{2}{c}{100-Shot-Panda}                \\ \hline
\multicolumn{1}{c}{D} & \multicolumn{1}{c}{G} & FID            & IS                     & FID            & IS                     & FID             & IS                     \\ \hline
                      &                       & 44.66          & 2.36$\pm$0.07          & 46.90          & 1.29$\pm$0.01          & 10.42           & 1.00$\pm$0.00          \\
  \CheckmarkBold      &                       & 43.43          & 2.44$\pm$0.02          & 40.91          & 1.48$\pm$0.03          & 9.97            & 1.02$\pm$0.00          \\
  \CheckmarkBold      & \CheckmarkBold        & \textbf{33.22} & \textbf{2.66}$\pm$0.06 & \textbf{39.41} & \textbf{1.53}$\pm$0.02 & \textbf{9.53}   & \textbf{1.17}$\pm$0.01 \\ \hline
\multicolumn{8}{c}{\textbf{512*512 Datasets}}                                                                                                                                         \\ \hline
\multicolumn{2}{l}{Augment?}                  & \multicolumn{2}{c}{Anime Face}          & \multicolumn{2}{c}{Art Paintings}       & \multicolumn{2}{c}{moongate}             \\ \hline
\multicolumn{1}{c}{D} & \multicolumn{1}{c}{G} & FID            & IS                     & FID            & IS                     & FID             & IS                     \\ \hline
                      &                       & 75.63          & 1.93$\pm$0.02          & 49.69          & 3.68$\pm$0.07          & 148.34          & 3.05$\pm$0.05          \\
  \CheckmarkBold      &                       & 56.40          & 1.90$\pm$0.01          & 47.45          & 3.70$\pm$0.12          & 116.06          & 2.98$\pm$0.06          \\
  \CheckmarkBold      & \CheckmarkBold        & \textbf{52.75} & \textbf{2.32}$\pm$0.05 & \textbf{44.70} & \textbf{3.77}$\pm$0.08 & \textbf{114.43} & \textbf{3.98}$\pm$0.09 \\ \hline
\multicolumn{8}{c}{\textbf{1024*1024 Datasets}}                                                                                                                                       \\ \hline
\multicolumn{2}{l}{Augment?}                  & \multicolumn{2}{c}{Pokemon}             & \multicolumn{2}{c}{Skulls}              & \multicolumn{2}{c}{BrecaHAD}             \\ \hline
\multicolumn{1}{c}{D} & \multicolumn{1}{c}{G} & FID            & IS                     & FID            & IS                     & FID             & IS                     \\ \hline
                      &                       & 83.31          & 2.12$\pm$0.06          & 110.99         & 3.20$\pm$0.06          & 63.37           & 2.71$\pm$0.08          \\
  \CheckmarkBold      &                       & 66.83          & 2.26$\pm$0.02          & 99.83          & 3.87$\pm$0.06          & 59.31           & 2.90$\pm$0.03          \\
  \CheckmarkBold      & \CheckmarkBold        & \textbf{54.83} & \textbf{2.41}$\pm$0.05 & \textbf{94.88} & \textbf{4.37}$\pm$0.14 & \textbf{58.13}  & \textbf{2.99}$\pm$0.09 \\ \hline
\end{tabular}
}
\end{table}

\textbf{Quantitative Results.}
We compare our model with transfer learning based and training from scratch methods on 256*256 datasets.
The FID results are presented in Tab.~\ref{Res256} and the IS results are given in Tab.~\ref{IS256}, from which we can observe that our proposed approach achieves state-of-the-art performance in $7$ of $9$ datasets.
Especially when compared with the baseline without semantic augmentation, our approach alleviates the performance deterioration.
\begin{table*}
\centering
\caption{\textbf{FID({$\downarrow$}) comparison on 256*256 datasets.} The numbers in parentheses after the datasets name denote the total number of training images.
$^\dagger$ Results are quoted from the results reported in~\cite{liu2021towards} and~\cite{DiffAug}.
$^\ddagger$ Results are re-implemented under the same setting with the official code released by the author.
The best results and the second-ranked results are \textbf{bold} and \underline{underlined}, respectively.}
\label{Res256}
\resizebox{\textwidth}{!}{
\begin{tabular}{cccc|ccccccc}
\hline
\multirow{2}{*}{Methods} & \multirow{2}{*}{Pre-Training?} & \multicolumn{2}{c|}{\textbf{Animal Face}} & \multicolumn{7}{c}{\textbf{100-Shot}}                                                                                    \\
                         &                                & Dog(389)    & Cat(160)          & Panda         & Grumpy-cat & Obama          & Wuzhen         & Bridge         & Medici         & Temple         \\ \hline
Scale/shift{$^\dagger$}~\cite{noguchi2019image}               & \CheckmarkBold       & 83.04        & 54.83             & 21.38         & 34.20      & 50.72          & -              & -              & -              & -              \\
FreezeD{$^\dagger$}~\cite{mo2020freeze}                       & \CheckmarkBold       & 70.46        & 47.70             & 17.95         & 31.22      & 41.87          & -              & -              & -              & -              \\
MineGAN{$^\dagger$}~\cite{wang2020minegan}                    & \CheckmarkBold       & 93.03        & 54.45             & 14.84         & 34.54      & 50.63          & -              & -              & -              & -              \\
TransferGAN{$^\dagger$} ~\cite{wang2018transferring}          & \CheckmarkBold       & 82.38        & 52.61             & 23.20         & 34.06      & 48.73          & -              & -              & -              & -              \\
TransferGAN+DiffAug{$^\dagger$}~\cite{DiffAug}            & \CheckmarkBold       & 65.57        & 49.10             & 17.12         & 29.77      & 39.85          & -              & -              & -              & -              \\
StyleGAN finetune{$^\dagger$}~\cite{liu2021towards}           & \CheckmarkBold       & 61.03        & 46.07             & 14.50         & 29.34      & \textbf{35.75} & -              & -              & -              & -              \\
StyleGAN2{$^\ddagger$}~\cite{karras2020analyzing}             & \XSolidBrush         & 113.86       & 79.04             & 18.05         & 35.00      & 69.01          & 135.40         & 116.40         & 85.73          & 73.35          \\
StyleGAN2+DiffAug{$^\ddagger$}~\cite{DiffAug}             & \XSolidBrush         & 61.34        & 41.84             & 11.52         & 26.89      & 48.85          & 122.44         & 49.97          &\underline{42.63}& 50.73         \\
ADA{$^\ddagger$}~\cite{karras2020training}                    & \XSolidBrush         & 55.48        & 37.95             & 14.17         & 43.80      & 51.25          & 92.81          & 72.07          & 44.21          & 49.72          \\
FastGAN{$^\ddagger$}~\cite{liu2021towards}                    & \XSolidBrush         &\textbf{50.66}&\underline{35.11}  & \underline{10.03} & \underline{26.65} & 41.05      & \underline{71.44}        & \underline{48.03}          & 47.77          & \underline{44.65}          \\ \hline
Baseline                                                      & \XSolidBrush         & 72.12        & 44.66             & 10.42         & 29.32      & 46.90          & 79.85          & 74.91          & 50.66          & 47.42          \\
Ours ASA-GAN                                                  & \XSolidBrush         &\underline{51.65} &\textbf{33.22} & \textbf{9.53} & \textbf{26.57}    & \underline{39.41} & \textbf{69.48} & \textbf{47.02} & \textbf{41.10} & \textbf{42.41} \\ \hline
\end{tabular}
}
\end{table*}

\begin{table*}
\centering
\caption{\textbf{IS({$\uparrow$}) comparison on 256*256 datasets.} All the results are re-implemented under the same setting for fair comparisons. The best results and the second-ranked results are bolded and underlined, respectively.}
\label{IS256}
\resizebox{\textwidth}{!}{
\begin{tabular}{ccc|ccccccc}
\hline
\multirow{2}{*}{Methods} & \multicolumn{2}{c|}{\textbf{Animal Face}}            & \multicolumn{7}{c}{\textbf{100-shot}}                                                                                                                                                      \\ \cline{2-10}
                                                    & Dog(389)                  & Cat(160)                  & Panda                    & Grumpy-cat         & obama              & \multicolumn{1}{l}{wuzhen} & \multicolumn{1}{l}{bridge} & \multicolumn{1}{l}{medici} & \multicolumn{1}{l}{templet} \\
StyleGAN2~\cite{karras2020analyzing}                & 7.28$\pm$0.32             & \underline{2.37}$\pm$0.08 & \underline{1.02}$\pm$0.00          & 1.31$\pm$0.01          & 1.37$\pm$0.04          & 1.93$\pm$0.04              & 1.60$\pm$0.02                  & 1.72$\pm$0.04            & 1.54$\pm$0.06                   \\
DiffAug~\cite{DiffAug}                          & \textbf{8.21}$\pm$0.30    & 2.02$\pm$0.04             & 1.00$\pm$0.00            & 1.28$\pm$0.01             & 1.29$\pm$0.02             & 1.92$\pm$0.04                  & 1.67$\pm$0.02                  & 1.78$\pm$0.03            & \underline{1.75}$\pm$0.02       \\
ADA~\cite{karras2020training}                       & \underline{8.13}$\pm$0.36 & 2.33$\pm$0.06             & 1.00$\pm$0.00            & 1.10$\pm$0.00             & 1.37$\pm$0.02             & \underline{2.01}$\pm$0.02      & \underline{1.91}$\pm$0.03      & \textbf{2.17}$\pm$0.03   & \textbf{1.98}$\pm$0.03          \\
FastGAN~\cite{liu2021towards}                       & 7.65$\pm$0.24             & 2.28$\pm$0.03             & \underline{1.02}$\pm$0.00& \underline{1.35}$\pm$0.02 & \underline{1.45}$\pm$0.02 & 1.94$\pm$0.03                  & 1.49$\pm$0.02                  & 1.99$\pm$0.03            & 1.61$\pm$0.02                   \\ \hline
Baseline                                            & 7.15$\pm$0.17             & 2.36$\pm$0.07             & 1.00$\pm$0.00            & 1.33$\pm$0.01             & 1.29$\pm$0.01             & 2.00$\pm$0.04                  & 1.82$\pm$0.02                  & 2.05$\pm$0.02            & 1.74$\pm$0.04                   \\
Ours ASA-GAN                                        & 7.45$\pm$0.20             & \textbf{2.66}$\pm$0.06    & \textbf{1.17}$\pm$0.01   & \textbf{1.36}$\pm$0.01    & \textbf{1.53}$\pm$0.02    & \textbf{2.18}$\pm$0.05         & \textbf{2.34}$\pm$0.06         & \underline{2.08}$\pm$0.02& \underline{1.75}$\pm$0.03       \\ \hline
\end{tabular}
}
\end{table*}

\begin{figure}
  \centering
  \includegraphics[width=\linewidth]{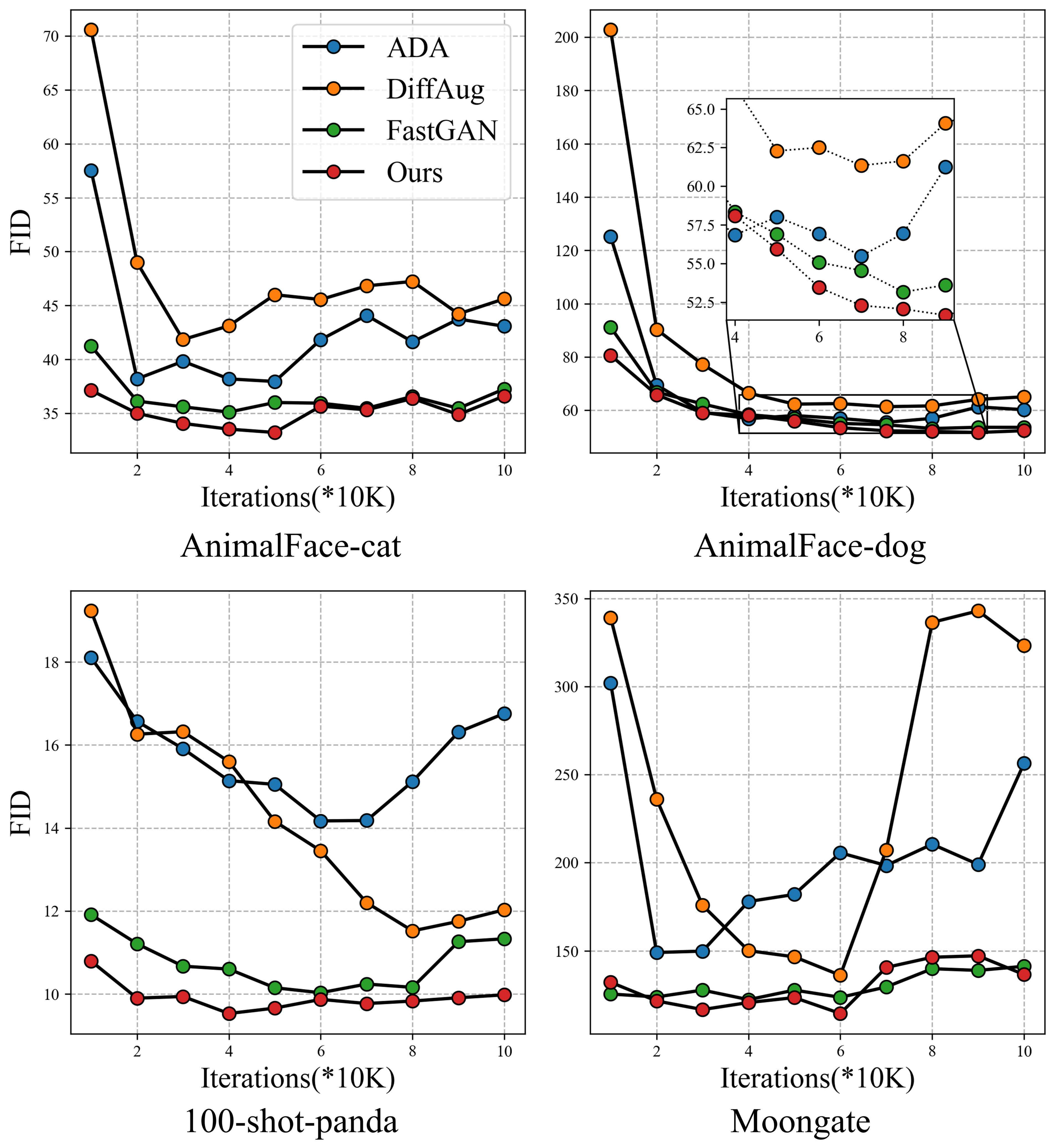}
  \caption{\textbf{Comparison on FID dynamics} of our model and DiffAug~\cite{DiffAug}, ADA~\cite{karras2020training}, FastGAN~\cite{liu2021towards}. Our model converges faster and stabilizes the training process, alleviating the overfitting issue.}
  \label{Dynamics}
  \vspace{-1mm}
\end{figure}

We further experiment on higher resolution datasets, including 512*512 and 1024*1024. As summarized in Tab.~\ref{Res512} and Tab.~\ref{Res1024},
the results substantiate that our model improves the generalization performance when data is limited.
Furthermore, we plot the FID curves during the model training to understand the convergence dynamics in Fig.~\ref{Dynamics}.
We can observe from the figure that our method stabilizes the training process and alleviates the overfitting.
Meanwhile, the obtained lower FID scores in the early stage of the training demonstrate that our method converges faster, suggesting the superiority and efficiency of our model.

\begin{table*}
\centering
\caption{\textbf{FID({$\downarrow$}) and IS({$\uparrow$}) comparison on 512*512 datasets.}
 The numbers in parentheses after the datasets name denote the total number of training images.
 All the results are re-implemented under the same setting for fair comparisons.}
\label{Res512}
\resizebox{\textwidth}{!}{
\begin{tabular}{ccccccccccc}
\hline
\multirow{2}{*}{Methods} & \multicolumn{2}{c}{Anime Face(120)} & \multicolumn{2}{c}{Art Paintings(1000)} & \multicolumn{2}{c}{Moongate(136)} & \multicolumn{2}{c}{Flat-colored(36)} &  \multicolumn{2}{c}{Fauvism-still-life(124)} \\ \cline{2-11}
& FID({$\downarrow$})        & IS({$\uparrow$})                  & FID({$\downarrow$})                & IS({$\uparrow$})                   & FID({$\downarrow$})                & IS({$\uparrow$})           & FID({$\downarrow$})            & IS({$\uparrow$})                & FID({$\downarrow$})                & IS({$\uparrow$})                   \\ \hline
StyleGAN2~\cite{karras2020analyzing}         & 183.44            & 1.37$\pm$0.01                      & 100.35           & 3.17$\pm$0.05            & 288.25             & 3.88$\pm$0.11             & 285.61            & 2.20$\pm$0.03           & \textbf{181.91}     & 2.33$\pm$0.04             \\
DiffAug~\cite{DiffAug}                   & 135.85            & 1.19$\pm$0.01                      & 49.25            & 3.38$\pm$0.05            & 136.12             & 2.65$\pm$0.07             & 340.14            & 1.48$\pm$0.01           & 223.58              & 3.10$\pm$0.10             \\
ADA~\cite{karras2020training}                & 59.67             & 1.92$\pm$0.03                      & \underline{46.38}& 3.64$\pm$0.12            & 149.06             & \textbf{4.11}$\pm$0.19    & 248.46            & 3.77$\pm$0.09           & 201.99              & 2.93$\pm$0.07             \\
FastGAN~\cite{liu2021towards}                & \underline{54.10} & \underline{2.04}$\pm$0.02          & 46.99            & \textbf{4.44}$\pm$0.06   & \underline{122.29} & 3.24$\pm$0.08             & \underline{240.24}&\underline{4.62}$\pm$0.09& \underline{182.14}  & \textbf{3.36}$\pm$0.10    \\ \hline
Baseline                                     & 75.63             & 1.93$\pm$0.02                      & 49.69            & 3.68$\pm$0.07            & 148.34             & 3.05$\pm$0.05             & 245.45            & 4.10$\pm$0.13           & 184.97              & 2.89$\pm$0.08             \\
Ours ASA-GAN                                 & \textbf{52.75}    & \textbf{2.32}$\pm$0.05             & \textbf{44.70}   & \underline{3.77}$\pm$0.08& \textbf{114.43}    & \underline{3.98}$\pm$0.09 & \textbf{227.89}   & \textbf{4.68}$\pm$0.12  & 182.92              & \underline{2.94}$\pm$0.06 \\ \hline
\end{tabular}
}
\end{table*}

\textbf{Qualitative Results.}
We give the qualitative comparative results in Fig.~\ref{VisAll}. The merits of our method become more evident from the synthesized images in the figure.
\revise{We could tell that, the synthesized images of compared baselines (\emph{i.e.,} StyleGAN2~\cite{karras2020analyzing} and current state-of-the-art few-shot method FastGAN~\cite{liu2021towards}) contain more obvious artifacts than ours.
For instance, StyleGAN2 synthesis human faces and anime faces with unnatural facial characteristics such as eyes.
FastGAN performs better but still have some artifacts and distortions, such as the teeth of skulls, the symmetry and rationality of animal and human faces.}
In comparison, the images synthesized by our model are more reasonable and realistic, which further validate the effectiveness of our method.

\begin{figure}
\centering
\includegraphics[width=\linewidth]{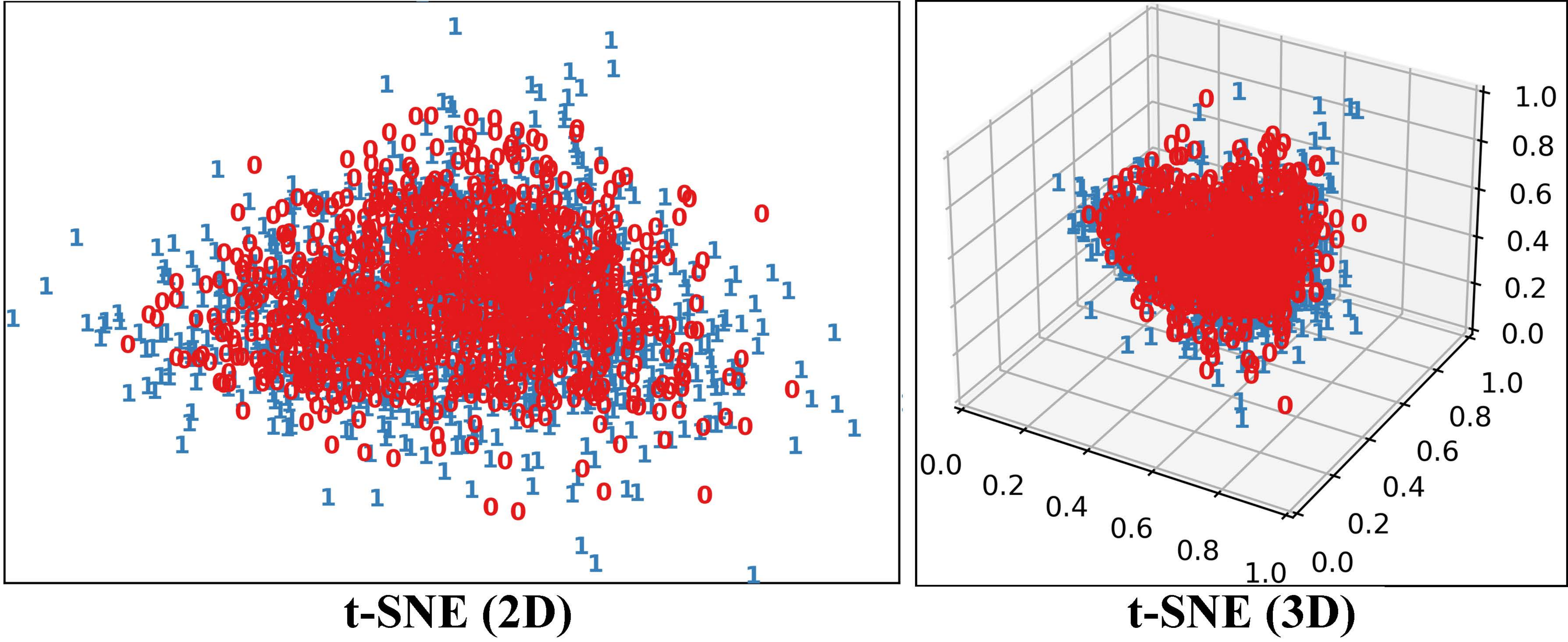}
\caption{
    \textbf{t-SNE~\cite{van2008visualizing} visualization on Art Painting}.
    $1$ and $0$ represent the real and synthesized images, respectively.
    We use the whole training images as the real distribution and randomly generate the same number of images as the training images with our generator.
}
\label{fig:tsneartpainting}
\end{figure}

\revise{In order to investigate the discrepancy between our synthesized and the real distribution, we use t-SNE~\cite{van2008visualizing} to visualize the two distribution.
Concretely, the whole training images are used as the real distribution, and we randomly synthesis the same number of images as the training images with our generator.
The synthesized images are viewed as the generated distribution.
Fig.~\ref{fig:tsneartpainting} present the visualization results.
Obviously, our synthesized distributions have large overlap with the referenced distributions, indicating that our method indeed learn to reproduce the original distribution.}

\begin{table*}
\centering
\caption{\textbf{FID({$\downarrow$}) and IS({$\uparrow$}) comparison on 1024*1024 datasets.} All the results are re-implemented under the same setting.
The numbers in parentheses after the datasets name denote the total number of training images.}
\label{Res1024}
\resizebox{\textwidth}{!}{
\begin{tabular}{ccccccccccccc}
\hline
\multirow{2}{*}{Datasets} & \multicolumn{2}{c}{Pokemon(833)}             & \multicolumn{2}{c}{Skulls(97)}                   & \multicolumn{2}{c}{Shells(64)}                   & \multicolumn{2}{c}{BrecaHAD(162)}                & \multicolumn{2}{c}{MetFace(1336)}                & \multicolumn{2}{c}{Flowers(1000)}            \\ \cline{2-13}
                                                     & FID ({$\downarrow$})   & IS ({$\uparrow$})         & FID ({$\downarrow$})   & IS ({$\uparrow$})      & FID ({$\downarrow$})   & IS ({$\uparrow$})        & FID ({$\downarrow$})   & IS ({$\uparrow$})       & FID ({$\downarrow$})   & IS ({$\uparrow$})                   & FID ({$\downarrow$})   & IS ({$\uparrow$}) \\ \hline
StyleGAN2~\cite{karras2020analyzing}                 & 190.23                 & 2.10$\pm$0.03             & 235.54                 & 2.89$\pm$0.08          & 241.37                 & 3.01$\pm$0.10            & 174.07                 & 1.66$\pm$0.03           & 66.97                  & 2.90$\pm$0.03                       & 120.47                 & 2.69$\pm$0.06              \\
DiffAug~\cite{DiffAug}                           & 62.73                  & \textbf{2.59}$\pm$0.07    & 124.23                 & 2.67$\pm$0.08          & 151.94                 & \underline{3.14}$\pm$0.10& 93.71                  & 2.71$\pm$0.09           & \underline{27.45}      & \underline{3.29}$\pm$0.09           & 47.09                  & 3.14$\pm$0.13              \\
ADA~\cite{karras2020training}                        & 66.41                  & 1.41$\pm$0.02             & \underline{97.05}      & 2.71$\pm$0.08          & \underline{133.22}     & \textbf{3.77}$\pm$0.08   & 76.67                  & 2.81$\pm$0.09           & 27.62                  & \textbf{3.37}$\pm$0.07              & \textbf{27.36}         & 3.50$\pm$0.10              \\
FastGAN~\cite{liu2021towards}                        & 58.02                  & 2.36$\pm$0.04             & 101.40                 & 2.41$\pm$0.07          & 160.43                 & 2.84$\pm$0.06            & \underline{59.93}      & 2.74$\pm$0.09           & 29.75                  & 2.95$\pm$0.07                       & 30.43                  & \textbf{3.66}$\pm$0.10     \\ \hline
Baseline                                             & 83.31                  & 2.12$\pm$0.06             & 110.99                 & 3.20$\pm$0.06          & 160.91                 & 2.93$\pm$0.06            & 63.37                  & 2.71$\pm$0.08           & 29.96                  &  2.96$\pm$0.09                      & 47.37                  & 3.54$\pm$0.08              \\
Ours ASA-GAN                                         & \textbf{54.83}         & \underline{2.41}$\pm$0.05 & \textbf{94.88}         & \textbf{4.37}$\pm$0.14 & \textbf{123.09}        & \textbf{3.77}$\pm$0.13   & \textbf{58.13}         & \textbf{2.99}$\pm$0.09  & \textbf{27.35}         & 3.04$\pm$0.05                       & \underline{29.76}      & \underline{3.64}$\pm$0.10  \\ \hline
\end{tabular}
}
\end{table*}

\begin{figure}
  \vspace{-4mm}
  \centering
  \includegraphics[width=.6\linewidth]{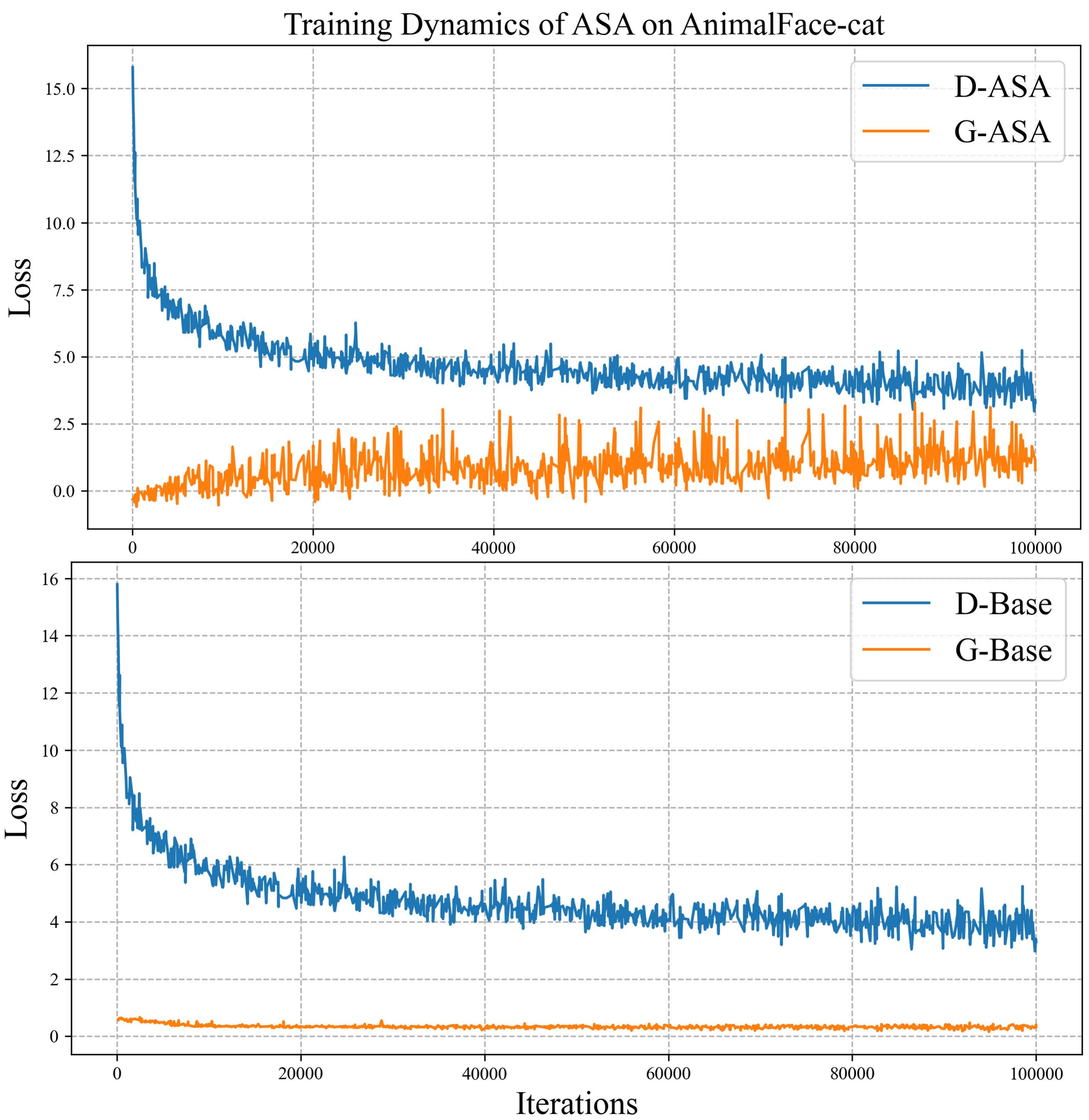}
  \caption{\textbf{Analysis of training loss with and without semantic augmentation.}
  Our semantic augmentation reduce the gap between the discriminator`s loss and the generator`s loss, leading to a better Nash equilibrium.}
  \label{LossCurve}
  \vspace{-1mm}
\end{figure}

\textbf{Training with More Samples.}
We evaluate if the proposed semantic augmentation technique still works for more comprehensive investigations when training data is sufficient.
We choose three datasets with different resolutions, namely CelebA~\cite{liu2015faceattributes} with 256*256 resolution, AFHQ~\cite{choi2020stargan} with 512*512 resolution, and FFHQ~\cite{karras2019style} with 1024*1024 resolution.
Tab.~\ref{Moresamples} presents the quantitative results.
Obviously, our method could also promote the synthesis performance when more samples are given.
Such observation demonstrates that our semantic augmentation technique indeed improve the data efficiency of models.
Additionally, the model parameters and training time are shown in Tab.~\ref{Moresamples}, from which we could tell that the proposed methods introduces negligible parameters, facilitating efficient training.
\begin{figure}
  \centering
  \includegraphics[width=\linewidth]{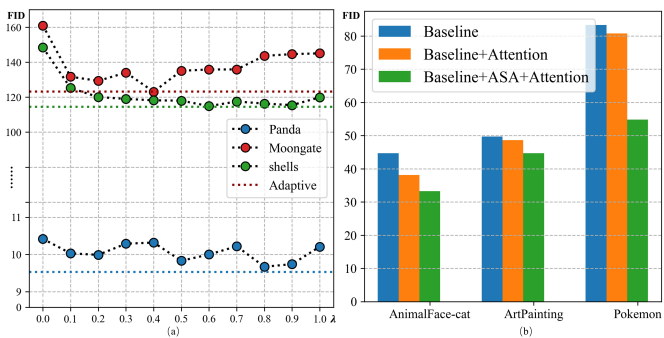}
  \caption{\textbf{(a) Impact of augmenting strengths. (b) Ablation Study on attention and adversarial semantic augmentation.}}
  \label{Fig:Lambda}
  \vspace{-2mm}
\end{figure}

\begin{table*}
\centering
\caption{\textbf{FID({$\downarrow$}) and IS({$\uparrow$}) comparison on datasets with more images.} All the results are re-implemented under the same setting.}
\label{Moresamples}
\resizebox{\textwidth}{!}{
\begin{tabular}{clcccc|cccccc|cccc|c}
\hline
\multirow{2}{*}{Datasets} & \multicolumn{5}{c|}{\textbf{CelebA(256)}}                                          & \multicolumn{6}{c|}{\textbf{AFHQ(512)}}                                                                                 & \multicolumn{4}{c}{\textbf{FFHQ(1024)}}                                \\ \cline{3-17}
                          & \multicolumn{3}{c}{5K}              & \multicolumn{2}{c|}{10K}            & \multicolumn{2}{c}{Cat(5153)}      & \multicolumn{2}{c}{Dog(4739)}        & \multicolumn{2}{c|}{Wild(4738)}    & \multicolumn{2}{c}{5K}              & \multicolumn{3}{c}{10K} \\ \hline
\multicolumn{1}{l}{       }                      & Params             & FID                 & IS                           & FID              & IS                     & FID           & IS                 & FID            & IS                  & FID           & IS                 & FID            & IS                 & FID      & IS                   & Training Time            \\ \hline
Baseline                                         & 40.72M             & 28.13               & 2.44$\pm$0.09                & 32.74            & 2.47$\pm$0.05              & 25.49             & 1.21$\pm$0.01               & 46.99               & 7.80$\pm$0.28              & 18.24            & 4.10$\pm$0.11            & 41.50             & 3.54$\pm$0.12          & 48.34                & 3.47$\pm$0.15             & 15.43 hours\\
Ours ASA-GAN                                      & 40.73M             & 22.75               & 2.50$\pm$0.04                & 22.32            & \underline{2.67}$\pm$0.07  & 10.21             & 1.98$\pm$0.04               & 25.67               & 8.09$\pm$0.25              & 7.82             & 4.90$\pm$0.08            & 29.10             & 3.40$\pm$0.13          & 22.49                & 3.68$\pm$0.08             & 15.63 hours\\ \hline
\end{tabular}
}
\end{table*}

\subsection{More Discussions and Visualization}

\textbf{Impact of the Augmentation Strength.}
We perform sensitivity analysis on the strength $\lambda$ of the semantic augmentation.
The result in Fig.~\ref{Fig:Lambda} (a) suggests that setting fixed $\lambda$ achieves poorer performance compared with adjusting $\lambda$ dynamically as training progresses.
This observation is reasonable since the feature extraction of the model is relatively weak in the early stage of training, we should set a lower value of $\lambda$.
As the training progresses, we increase the intensity of augmentation gradually.

\textbf{Ablation Study.} To evaluate the effects of attention block and our semantic augmentation technique, we conduct ablation studies on three conditions: 1) baseline without attention block and ASA; 2) with attention only; 3) with both attention and semantic augmentation.
As shown in Fig.~\ref{Fig:Lambda} (b), both of the two modules contribute to the performance of our model, among which our semantic augmentation improves more.
Combining with the results of augmenting images at different levels in Tab.~\ref{Table:Augment}, our method achieves favorable performance by utilizing both attention block and semantic augmentation.

\textbf{Loss of Discriminator and Generator.} We plot the predictions of both models with and without semantic augmentation in Fig.~\ref{LossCurve}.
As illustrated in the figure, the discriminator overfits easily and provides limited meaningful feedback to the generator.
The constant loss curve of the generator indicates that the discriminator memorizes the limited training images and fails to generalize.
By contrast, the discriminator provides more effective guidelines to the generator when augmentation is applied.
Such observation substantiates that our ASA: 1) mitigates the overfitting of $D$ and 2) facilitates $D$ and $G$ to reach Nash equilibrium.

\begin{figure}
  \vspace{-2mm}
  \centering
  \includegraphics[width=\linewidth]{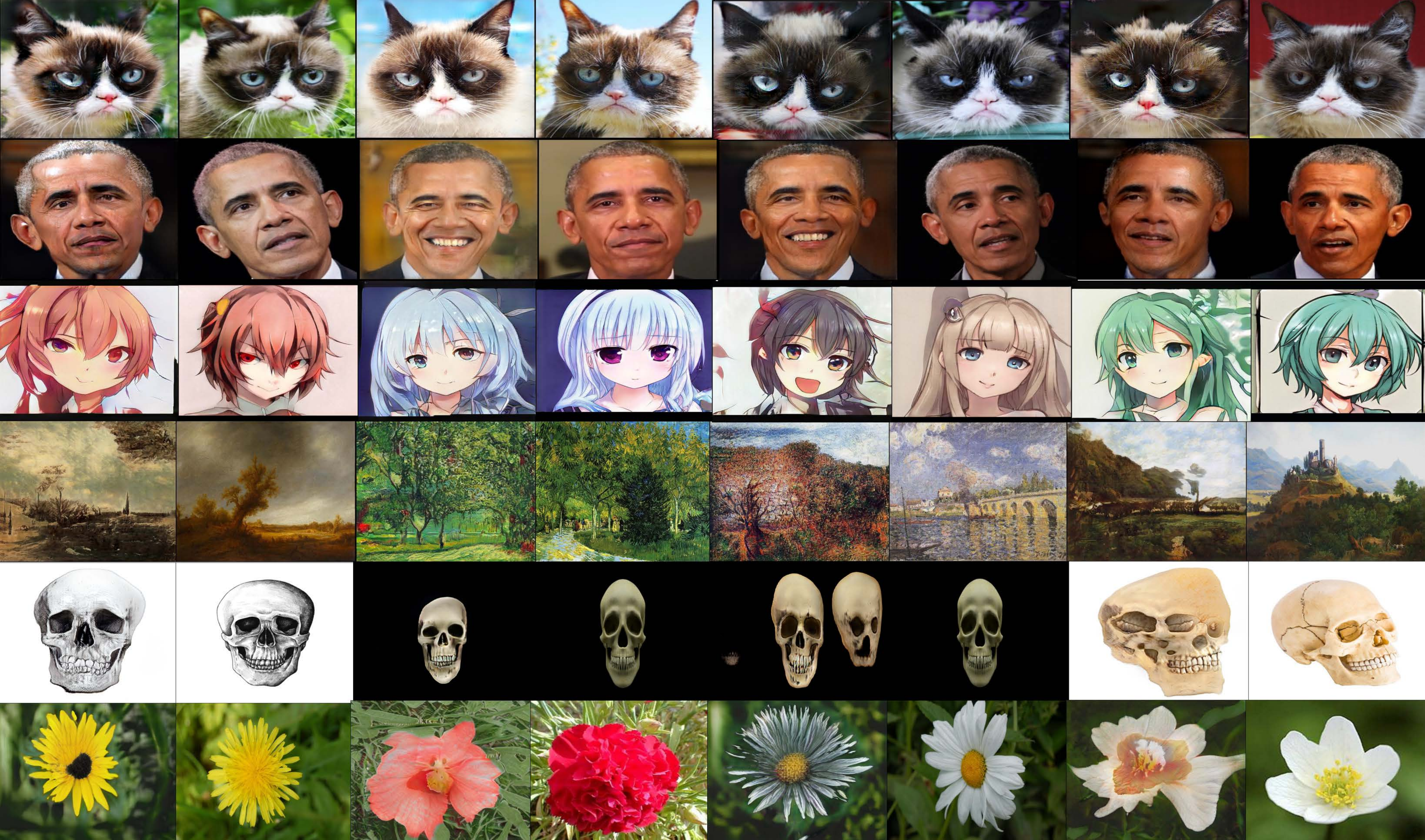}
  \caption{\textbf{Nearest real images to the generated one.} For each pair of images, the left images are generated our model and the right images are  real images found from training sets.
  Zoom in for a better view.
  }
  \label{NearestSamples}
  \vspace{-1mm}
\end{figure}

\textbf{Nearest Samples.} To demonstrate that our model can perceive the semantic features of the real images, we find the real images that are closest to the generated images by computing the LPIPS distance~\cite{zhang2018unreasonable}.
Fig.~\ref{NearestSamples} shows the paired neighbor images.
For each pair of images, the left is the generated image and the right is the most similar image found from the real images.
As can be seen from the figure, the semantic features of each pair of images have slight differences.
However, they vary significantly in the background, color, perspective, shape, and facial expressions, which demonstrates that our model learns to synthesize images instead of simply remembering the training images.

\begin{figure}
  \vspace{-2mm}
  \centering
  \includegraphics[width=\linewidth]{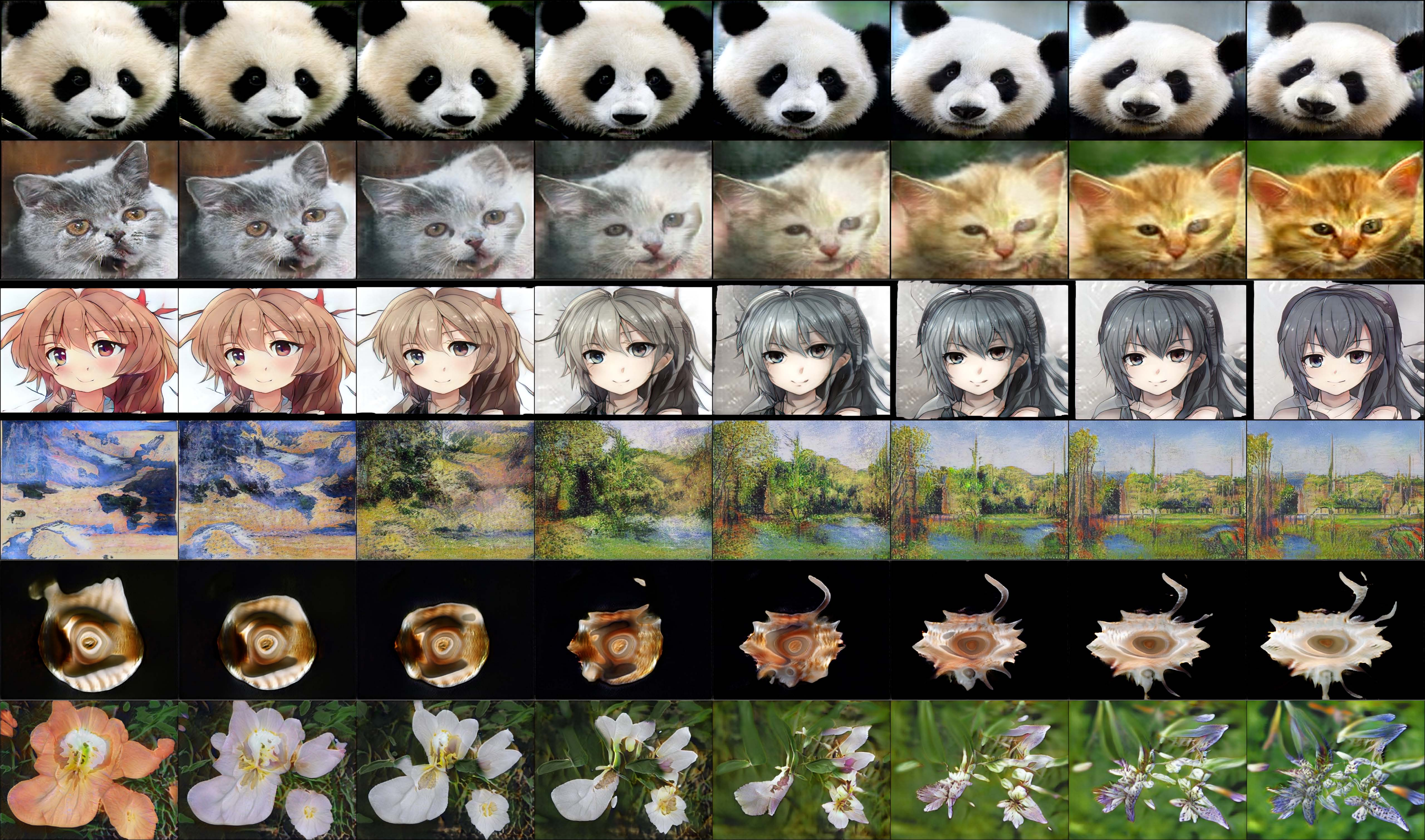}
  \caption{\textbf{Latent space interpolation results of our model.} The smooth interpolation results indicate that our model performs satisfactory performance and suffers from less overfitting.
  Zoom in for a better view.
  }
  \label{Backtracking}
  \vspace{-1mm}
\end{figure}

\textbf{Interpolation Results.} Despite synthesizing high-quality images, a well-trained GAN can also invert a real image into a latent code in the style space.
In this way, one can manipulate the content of the images by modifying the latent code.
The inversion performance is another way for evaluating the GAN model.
Consequently, we further test the efficacy of our model by performing latent space interpolation in Fig.~\ref{Backtracking}.
The smooth interpolation results in Fig.~\ref{Backtracking} indicate that our model, although trained on limited data, still achieves satisfactory performance and suffers less from overfitting.

\section{Limitations and Conclusion}

\revise{\textbf{Limitations.}
Despite achieving substantial performance gains on various datasets with limited training data, there are several limitations of our proposed methods for further improvement.
First, when given imbalanced or even long-tailed low-data datasets, the proposed method might fail to capture the overall data distribution.
Second, our proposed method struggles in producing plausible images when the training data contains various contents, \emph{i.e.,} only hundreds of images but their contents vary widely.
Moreover, our proposed method inherits the entanglement of GANs' latent space, making it challenging to edit various attributes of generated images.
Potential ways to address these limitations including 1) designing cost sensitive losses for imbalanced training, 2) injecting class-condition into the network during training, and 3) employing inversion techniques to disentangle the latent space.
In our future research, we plan pour more efforts to these problems.
}

\textbf{Conclusion.}
In this paper, we propose a novel adversarial semantic augmentation approach (ASA) for training GANs under limited data.
Unlike traditional data augmentation methods that perform augmentation at the image level, our model augments the training images at the semantic level.
We derive an upper bound of the adversarial loss by optimizing which the training samples are implicitly augmented.
The proposed semantic augmentation makes no change to the original data distribution and introduces negligible computational costs.
Both quantitative and qualitative results demonstrate the effectiveness of our model on various data regimes.
%

\section*{Acknowledgment}
This work is supported in part by the Natural Science Foundation of China under Grant 62476087, in part by the National Key Research and Development Program of China under Grant 2022YFB3203500.

\ifCLASSOPTIONcaptionsoff
  \newpage
\fi
\bibliographystyle{IEEEtran}
\bibliography{references}

\end{document}